\documentclass[10pt,a4paper,onecolumn]{article}
\usepackage{marginnote}
\usepackage{graphicx}
\usepackage{xcolor}
\usepackage{authblk,etoolbox}
\usepackage{titlesec}
\usepackage{calc}
\usepackage{tikz}
\usepackage{hyperref}
\hypersetup{colorlinks,breaklinks=true,
            urlcolor=[rgb]{0.0, 0.5, 1.0},
            linkcolor=[rgb]{0.0, 0.5, 1.0}}
\usepackage{caption}
\usepackage{tcolorbox}
\usepackage{amssymb,amsmath}
\usepackage{ifxetex,ifluatex}
\usepackage{seqsplit}
\usepackage{xstring}

\usepackage{float}
\let\origfigure\figure
\let\endorigfigure\endfigure
\renewenvironment{figure}[1][2] {
    \expandafter\origfigure\expandafter[H]
} {
    \endorigfigure
}

\usepackage{fixltx2e} 
\usepackage[
  backend=biber,
]{biblatex}
\bibliography{paper.bib}

\let\textttOrig=\texttt
\def\texttt#1{\expandafter\textttOrig{\seqsplit{#1}}}
\renewcommand{\seqinsert}{\ifmmode
  \allowbreak
  \else\penalty6000\hspace{0pt plus 0.02em}\fi}

\makeatletter
\let\href@Orig=\href
\def\href@Urllike#1#2{\href@Orig{#1}{\begingroup
    \def\Url@String{#2}\Url@FormatString
    \endgroup}}
\def\href@Notdoi#1#2{\def\tempa{#1}\def\tempb{#2}%
  \ifx\tempa\tempb\relax\href@Urllike{#1}{#2}\else
  \href@Orig{#1}{#2}\fi}
\def\href#1#2{%
  \IfBeginWith{#1}{https://doi.org}%
  {\href@Urllike{#1}{#2}}{\href@Notdoi{#1}{#2}}}
\makeatother

\newlength{\cslhangindent}
\newlength{\csllabelwidth}
\newenvironment{CSLReferences}[3] 
 {
  \setlength{\parindent}{0pt}
  \ifodd #1 \everypar{\setlength{\hangindent}{\cslhangindent}}\ignorespaces\fi
  \ifnum #2 > 0
  \setlength{\parskip}{#2\baselineskip}
  \fi
 }%
 {}
\usepackage{calc}

\usepackage[top=3.5cm, bottom=3cm, right=1.5cm, left=1.0cm,
            headheight=2.2cm, reversemp, includemp, marginparwidth=4.5cm]{geometry}

\titleformat{\section}
  {\normalfont\sffamily\Large\bfseries}
  {}{0pt}{}
\titleformat{\subsection}
  {\normalfont\sffamily\large\bfseries}
  {}{0pt}{}
\titleformat{\subsubsection}
  {\normalfont\sffamily\bfseries}
  {}{0pt}{}
\titleformat*{\paragraph}
  {\sffamily\normalsize}

\usepackage{fancyhdr}
\makeatletter
\let\ps@plain\ps@fancy
\fancyheadoffset[L]{4.5cm}
\fancyfootoffset[L]{4.5cm}

\definecolor{linky}{rgb}{0.0, 0.5, 1.0}

\newtcolorbox{repobox}
   {colback=red, colframe=red!75!black,
     boxrule=0.5pt, arc=2pt, left=6pt, right=6pt, top=3pt, bottom=3pt}

\newcommand{\ExternalLink}{%
   \tikz[x=1.2ex, y=1.2ex, baseline=-0.05ex]{%
       \begin{scope}[x=1ex, y=1ex]
           \clip (-0.1,-0.1)
               --++ (-0, 1.2)
               --++ (0.6, 0)
               --++ (0, -0.6)
               --++ (0.6, 0)
               --++ (0, -1);
           \path[draw,
               line width = 0.5,
               rounded corners=0.5]
               (0,0) rectangle (1,1);
       \end{scope}
       \path[draw, line width = 0.5] (0.5, 0.5)
           -- (1, 1);
       \path[draw, line width = 0.5] (0.6, 1)
           -- (1, 1) -- (1, 0.6);
       }
   }

\patchcmd{\@maketitle}{center}{flushleft}{}{}
\patchcmd{\@maketitle}{center}{flushleft}{}{}
\patchcmd{\@maketitle}{\LARGE}{\LARGE\sffamily}{}{}
\def\maketitle{{%
  
  \AB@maketitle}}
\makeatletter
\renewcommand\AB@affilsepx{ \protect\Affilfont}
\renewcommand\AB@affilnote[1]{{\bfseries #1}\hspace{3pt}}
\renewcommand{\affil}[2][]%
   {\newaffiltrue\let\AB@blk@and\AB@pand
      \if\relax#1\relax\def\AB@note{\AB@thenote}\else\def\AB@note{#1}%
        \setcounter{Maxaffil}{0}\fi
        \begingroup
        \let\href=\href@Orig
        \let\texttt=\textttOrig
        \let\protect\@unexpandable@protect
        \def\thanks{\protect\thanks}\def\footnote{\protect\footnote}%
        \@temptokena=\expandafter{\AB@authors}%
        {\def\\{\protect\\\protect\Affilfont}\xdef\AB@temp{#2}}%
         \xdef\AB@authors{\the\@temptokena\AB@las\AB@au@str
         \protect\\[\affilsep]\protect\Affilfont\AB@temp}%
         \gdef\AB@las{}\gdef\AB@au@str{}%
        {\def\\{, \ignorespaces}\xdef\AB@temp{#2}}%
        \@temptokena=\expandafter{\AB@affillist}%
        \xdef\AB@affillist{\the\@temptokena \AB@affilsep
          \AB@affilnote{\AB@note}\protect\Affilfont\AB@temp}%
      \endgroup
       \let\AB@affilsep\AB@affilsepx
}
\makeatother

\renewcommand\Affilfont{\sffamily\small\mdseries}
\ifnum 0\ifxetex 1\fi\ifluatex 1\fi=0 
  \usepackage[T1]{fontenc}
  \usepackage[utf8]{inputenc}

\else 
  \ifxetex
    \usepackage{mathspec}
    \usepackage{fontspec}

  \else
    \usepackage{fontspec}
  \fi
  \defaultfontfeatures{Ligatures=TeX,Scale=MatchLowercase}

\fi
\IfFileExists{upquote.sty}{\usepackage{upquote}}{}
\IfFileExists{microtype.sty}{%
\usepackage{microtype}
\UseMicrotypeSet[protrusion]{basicmath} 
}{}

\usepackage{hyperref}
\hypersetup{unicode=true,
            pdftitle={ttta: Tools for Temporal Text Analysis},
            pdfborder={0 0 0},
            breaklinks=true}
\let\addcontentslineOrig=\addcontentsline
\def\addcontentsline#1#2#3{\bgroup
  \let\texttt=\textttOrig\addcontentslineOrig{#1}{#2}{#3}\egroup}
\let\markbothOrig\markboth
\def\markboth#1#2{\bgroup
  \let\texttt=\textttOrig\markbothOrig{#1}{#2}\egroup}
\let\markrightOrig\markright
\def\markright#1{\bgroup
  \let\texttt=\textttOrig\markrightOrig{#1}\egroup}

\usepackage{graphicx,grffile}
\makeatletter
\def\maxwidth{\ifdim\Gin@nat@width>\linewidth\linewidth\else\Gin@nat@width\fi}
\def\maxheight{\ifdim\Gin@nat@height>\textheight\textheight\else\Gin@nat@height\fi}
\makeatother
\setkeys{Gin}{width=\maxwidth,height=\maxheight,keepaspectratio}
\IfFileExists{parskip.sty}{%
\usepackage{parskip}
}{
\setlength{\parindent}{0pt}
\setlength{\parskip}{6pt plus 2pt minus 1pt}
}
\ifx\paragraph\undefined\else
\let\oldparagraph\paragraph
\renewcommand{\paragraph}[1]{\oldparagraph{#1}\mbox{}}
\fi
\ifx\subparagraph\undefined\else
\let\oldsubparagraph\subparagraph
\renewcommand{\subparagraph}[1]{\oldsubparagraph{#1}\mbox{}}
\fi

\title{ttta: Tools for Temporal Text Analysis}

        \author[1]{Kai-Robin Lange}
          \author[2]{Niklas Benner}
          \author[1]{Lars Grönberg}
          \author[3]{Aymane Hachcham}
          \author[3]{Imene Kolli}
          \author[1]{Jonas Rieger}
          \author[1]{Carsten Jentsch}
    
      \affil[1]{TU Dortmund University}
      \affil[2]{RWI - Leibniz Institute for Economic Research}
      \affil[3]{University of Zurich}
  \date{\vspace{-7ex}}

\begin{document}
\maketitle

\marginpar{

  \begin{flushleft}
  \sffamily\small

  {\bfseries DOI:} \href{https://doi.org/DOI unavailable}{\color{linky}{DOI unavailable}}

  \vspace{2mm}

  {\bfseries Software}
  \begin{itemize}
    \setlength\itemsep{0em}
    \item \href{N/A}{\color{linky}{Review}} \ExternalLink
    \item \href{NO_REPOSITORY}{\color{linky}{Repository}} \ExternalLink
    \item \href{DOI unavailable}{\color{linky}{Archive}} \ExternalLink
  \end{itemize}

  \vspace{2mm}

  \par\noindent\hrulefill\par

  \vspace{2mm}

  {\bfseries Editor:} \href{https://example.com}{Pending
Editor} \ExternalLink \\
  \vspace{1mm}
    {\bfseries Reviewers:}
  \begin{itemize}
  \setlength\itemsep{0em}
    \item \href{https://github.com/Pending Reviewers}{@Pending
Reviewers}
    \end{itemize}
    \vspace{2mm}

  {\bfseries Submitted:} N/A\\
  {\bfseries Published:} N/A

  \vspace{2mm}
  {\bfseries License}\\
  Authors of papers retain copyright and release the work under a Creative Commons Attribution 4.0 International License (\href{http://creativecommons.org/licenses/by/4.0/}{\color{linky}{CC BY 4.0}}).

  \end{flushleft}
}

\hypertarget{statement-of-need}{%
\section{Statement of need}\label{statement-of-need}}

Text data is inherently temporal. The meaning of words and phrases
changes over time, and the context in which they are used is constantly
evolving. This is not just true for social media data, where the
language used is rapidly influenced by current events, memes and trends,
but also for journalistic, economic or political text data. Most NLP
techniques however consider the corpus at hand to be homogenous in
regard to time. This is a simplification that can lead to biased
results, as the meaning of words and phrases can change over time. For
instance, running a classic Latent Dirichlet Allocation (Blei et al.,
2003) on a corpus that spans several years is not enough to capture
changes in the topics over time, but only portraits an ``average'' topic
distribution over the whole time span.

Researchers have developed a number of tools for analyzing text data
over time. However, these tools are often scattered across different
packages and libraries, making it difficult for researchers to use them
in a consistent and reproducible way.

The \texttt{ttta} package is supposed to serve as a collection of tools
for analyzing text data over time and can be accessed using \href{https://github.com/K-RLange/ttta}{its GitHub repository}.

\hypertarget{summary}{%
\section{Summary}\label{summary}}

In its current state, the \texttt{ttta} package includes diachronic
embeddings, dynamic topic modeling, and document scaling. These tools
can be used to track changes in language use, identify emerging topics,
and explore how the meaning of words and phrases has evolved over time.

\begin{figure}
\centering
\includegraphics{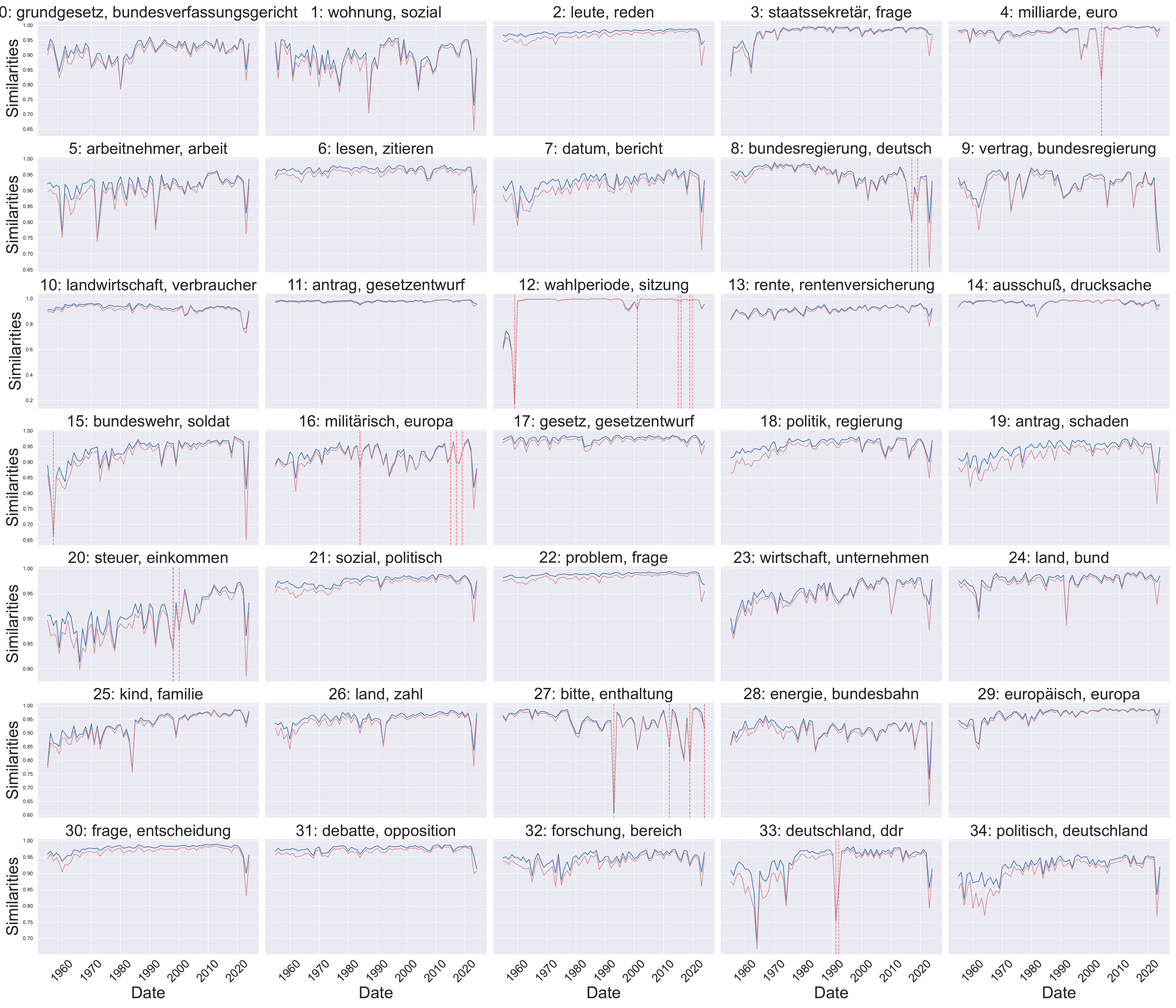}
\caption{Changes observed by the Topical Changes Model in a corpus of
speeches held in the German Bundestag between 1949 and 2023. There is
one plot for each topic, with the topic's most defining words over the
time frame provided as a title for easier interpretation. Each plot
shows the stability of the topic over time (blue line) as well as a
threshold calculated with a monitoring procedure (orange line). A change
is detected, when the observed stability falls below the threshold,
indicated by red vertical lines.\label{fig:topical}}
\end{figure}

Our dynamic topic model approach is based on the model RollingLDA
(Rieger et al., 2021), which is a modification of the classic Latent
Dirichlet Allocation (Blei et al., 2003), that allows for the estimation
of topics over time using a rolling window approach. We additionally
implemented the model LDAPrototype (Rieger et al., 2020), serving as a
more consistent foundation for RollingLDA than a common LDA. With these
models, users can uncover and analyze topics of discussion in temporal
data sets and track even rapid changes, which other dynamic topic models
struggle with. This ability to track rapid changes in topics is further
used in the Topical Changes model put forth by Rieger et al. (2022) and
Lange et al. (2022) that identifies change points in the word topic
distribution of RollingLDA. \autoref{fig:topical} visualizes the changes
observed by the Topical Changes model in speeches from the German
Bundestag (Lange \& Jentsch, 2023), which can be analyzed further using
leave-one-out word impacts provided by the model or, as Lange et al.
(2025) proposed, by asking Large Language Models to interpret the change
and relate it to a possible narrative shift.

The first diachronic embedding model, originally introduced by Hamilton
et al. (2016), builds on the static word embedding model Word2Vec
(Mikolov et al., 2013). It enables the estimation of word embeddings
over time by aligning Word2Vec vector spaces across different time
chunks using a rotation matrix. The second diachronic embedding model is
based on the work of Hu et al. (2019), who leveraged BERT's contextual
language understanding to associate word usage in a sentence with a
specific word sense, thus enabling users to track shifts in word
meanings over time.

\begin{figure}
\centering
\includegraphics{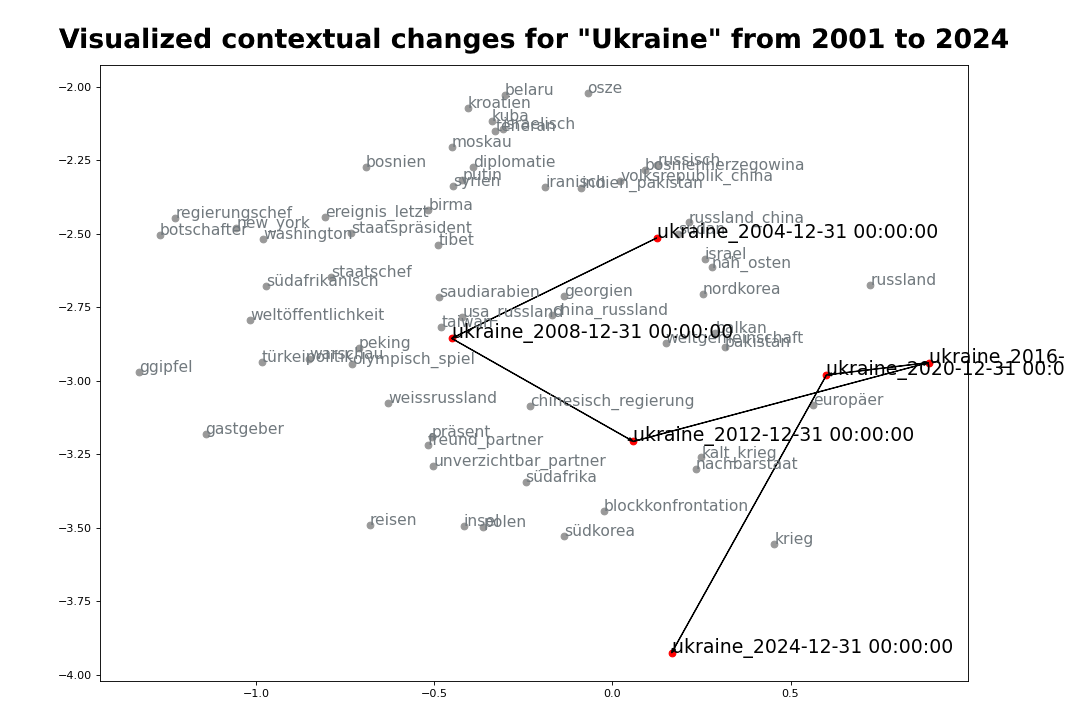}
\caption{Development of the diachronic embedding of the word ``ukraine''
from 2004 to 2024 in the German Bundestag. Along with the word itself,
its closest neighbors to visualize the target word's track across time.
The dimension of the embeddings has been lowered using
TSNE.\label{fig:ukraineplot}}
\end{figure}

An example of the evolution of the static diachronic embedding of the
word Ukraine in the German Bundestag from 2004 to 2024 is shown in
\autoref{fig:ukraineplot}. The plot shows the nearest neighbors of
Ukraine in the respective embedding spaces, enabling users to observe
the trajectory of the target word Ukraine across time chunks in a
low-dimensional representation. This visualization highlights the
potential of using diachronic embeddings for the interpretation and
detection of word context change, as it reflects Ukraine's contextual
shift from being closely associated with Russia and China, to aligning
more with Europe, and ultimately moving into a different context
centered on war.

The Poisson Reduced Rank model (Jentsch et al., 2020, 2021) is a
document scaling model, which uses a poisson-distribution based time
series analysis to model the word usage of different entities
(e.g.~parties when analyzing party manifestos). With this model, the
user are able to analyze, how the entities move in a latent space that
is generated using the word usage counts.

In future work, we plan to add more tools for analyzing temporal text
data, as we consider the current state of the package to only be the
beginning of development.

\hypertarget{acknowledgements}{%
\section{Acknowledgements}\label{acknowledgements}}

This paper is part of a project of the Dortmund Center for data-based
Media Analysis (DoCMA) at TU Dortmund University.

\hypertarget{references}{%
\section*{References}\label{references}}
\addcontentsline{toc}{section}{References}

\hypertarget{refs}{}
\begin{CSLReferences}{1}{0}
\leavevmode\hypertarget{ref-bleiLatentDirichletAllocation2003}{}%
Blei, D. M., Ng, A. Y., \& Jordan, M. I. (2003). Latent dirichlet
allocation. \emph{The Journal of Machine Learning Research}, \emph{3},
993--1022. \url{https://dl.acm.org/doi/10.5555/944919.944937}

\leavevmode\hypertarget{ref-Hamilton}{}%
Hamilton, W. L., Leskovec, J., \& Jurafsky, D. (2016). Diachronic word
embeddings reveal statistical laws of semantic change. \emph{Proceedings
of the 54th Annual Meeting of the Association for Computational
Linguistics (Volume 1: Long Papers)}.
\url{https://doi.org/10.18653/v1/P16-1141}

\leavevmode\hypertarget{ref-huDiachronicSenseModeling2019}{}%
Hu, R., Li, S., \& Liang, S. (2019). Diachronic {Sense Modeling} with
{Deep Contextualized Word Embeddings}: {An Ecological View}.
\emph{Proceedings of the 57th {Annual Meeting} of the {Association} for
{Computational Linguistics}}, 3899--3908.
\url{https://doi.org/10.18653/v1/P19-1379}

\leavevmode\hypertarget{ref-PRR1}{}%
Jentsch, C., Lee, E. R., \& Mammen, E. (2020). Time-dependent {P}oisson
reduced rank models for political text data analysis.
\emph{Computational Statistics \& Data Analysis}, \emph{142}.
\url{https://doi.org/10.1016/j.csda.2019.106813}

\leavevmode\hypertarget{ref-PRR2}{}%
Jentsch, C., Lee, E. R., \& Mammen, E. (2021). Poisson reduced-rank
models with an application to political text data. \emph{Biometrika},
\emph{108}(2). \url{https://doi.org/10.1093/biomet/asaa063}

\leavevmode\hypertarget{ref-SpeakGer}{}%
Lange, K.-R., \& Jentsch, C. (2023). {SpeakGer}: {A} meta-data
enriched speech corpus of {German} state and federal parliaments.
\emph{Proceedings of the 3nd Workshop on Computational Linguistics for the Political and Social Sciences}. \url{https://aclanthology.org/2023.cpss-1.3/}

\leavevmode\hypertarget{ref-zeitenwenden}{}%
Lange, K.-R., Rieger, J., Benner, N., \& Jentsch, C. (2022).
Zeitenwenden: {Detecting} changes in the {German} political discourse.
\emph{Proceedings of the 2nd Workshop on Computational Linguistics for
Political Text Analysis}.
\url{https://old.gscl.org/media/pages/arbeitskreise/cpss/cpss-2022/workshop-proceedings-2022/254133848-1662996909/cpss-2022-proceedings.pdf}

\leavevmode\hypertarget{ref-NarrativeShiftDetection}{}%
Lange, K.-R., Schmidt, T., Reccius, M., Roos, M., Müller, H., \&
Jentsch, C. (2025). Narrative shift detection: A hybrid approach of
dynamic topic models and large language models. \emph{ To appear in Proceedings of the
Text2Story'25 Workshop}.

\leavevmode\hypertarget{ref-mikolovEfficientEstimationWord2013}{}%
Mikolov, T., Chen, K., Corrado, G., \& Dean, J. (2013). \emph{Efficient
{Estimation} of {Word Representations} in {Vector Space}} (No.
arXiv:1301.3781). \url{http://arxiv.org/abs/1301.3781}

\leavevmode\hypertarget{ref-RollingLDA}{}%
Rieger, J., Jentsch, C., \& Rahnenführer, J. (2021). {RollingLDA}: {A}n
update algorithm of {L}atent {D}irichlet {A}llocation to construct
consistent time series from textual data. \emph{Findings Proceedings of
the 2021 EMNLP-Conference}.
\url{https://doi.org/10.18653/v1/2021.findings-emnlp.201}

\leavevmode\hypertarget{ref-TopicalChanges}{}%
Rieger, J., Lange, K.-R., Flossdorf, J., \& Jentsch, C. (2022). Dynamic
change detection in topics based on rolling {LDAs}. \emph{Proceedings of
the Text2Story'22 Workshop}.
\url{https://ceur-ws.org/Vol-3117/paper1.pdf}

\leavevmode\hypertarget{ref-riegerImprovingLatentDirichlet2020}{}%
Rieger, J., Rahnenführer, J., \& Jentsch, C. (2020). Improving {Latent
Dirichlet Allocation}: {On Reliability} of the {Novel Method
LDAPrototype}. \emph{Natural {Language Processing} and {Information
Systems}}, 118--125. \url{https://doi.org/10.1007/978-3-030-51310-8_11}

\end{CSLReferences}

\end{document}